\documentclass[sigconf]{acmart}

\usepackage{cleveref}

\usepackage{booktabs} 
\usepackage{multirow} 
\usepackage{array}    
\usepackage{geometry} 
\AtBeginDocument{%
  }

\acmConference[SIGIR '24]{The 47th International ACM SIGIR Conference on Research and Development in Information Retrieval}{July 14--18, 2024}{Washington, DC}

\begin{document}

\title{Large Language Model Informed Patent Image Retrieval}


\author{Hao-Cheng Lo}
\orcid{0009-0005-4176-4861}
\affiliation{%
  \institution{National Taiwan University}
  \city{Taipei}
  \country{Taiwan}
}
\affiliation{%
  \institution{JCIPRNET}
  \city{Irvine}
  \state{California}
  \country{USA}
}
\email{austenpsy@gmail.com}

\author{Jung-Mei Chu}
\affiliation{%
  \institution{National Taiwan University}
  \city{Taipei}
  \country{Taiwan}
}
\affiliation{%
  \institution{JCIPRNET}
  \city{Irvine}
  \state{California}
  \country{USA}
}
\email{d09944017@csie.ntu.edu.tw}

\author{Jieh Hsiang}
\affiliation{%
  \institution{National Taiwan University}
  \city{Taipei}
  \country{Taiwan}
}
\email{hsiang@csie.ntu.edu.tw}

\author{Chun-Chieh Cho}
\affiliation{%
  \institution{JCIPRNET}
  \city{Irvine}
  \state{California}
  \country{USA}
}
\email{jeff@jcipgroup.com}

\renewcommand{\shortauthors}{Lo et al.}

\begin{abstract}
  In patent prosecution, image-based retrieval systems for identifying similarities between current patent images and prior art are pivotal to ensure the novelty and non-obviousness of patent applications. Despite their growing popularity in recent years, existing attempts, while effective at recognizing images within the same patent, fail to deliver practical value due to their limited generalizability in retrieving relevant prior art. Moreover, this task inherently involves the challenges posed by the abstract visual features of patent images, the skewed distribution of image classifications, and the semantic information of image descriptions. Therefore, we propose a language-informed, distribution-aware multimodal approach to patent image feature learning, which enriches the semantic understanding of patent image by integrating Large Language Models and improves the performance of underrepresented classes with our proposed distribution-aware contrastive losses. Extensive experiments on DeepPatent2 dataset show that our proposed method achieves state-of-the-art or comparable performance in image-based patent retrieval with mAP $+53.3\%$, Recall@10 $+41.8\%$, and MRR@10 $+51.9\%$. Furthermore, through an in-depth user analysis, we explore our model in aiding patent professionals in their image retrieval efforts, highlighting the model's real-world applicability and effectiveness.
\end{abstract}



\begin{teaserfigure}
  \includegraphics[width=\textwidth]{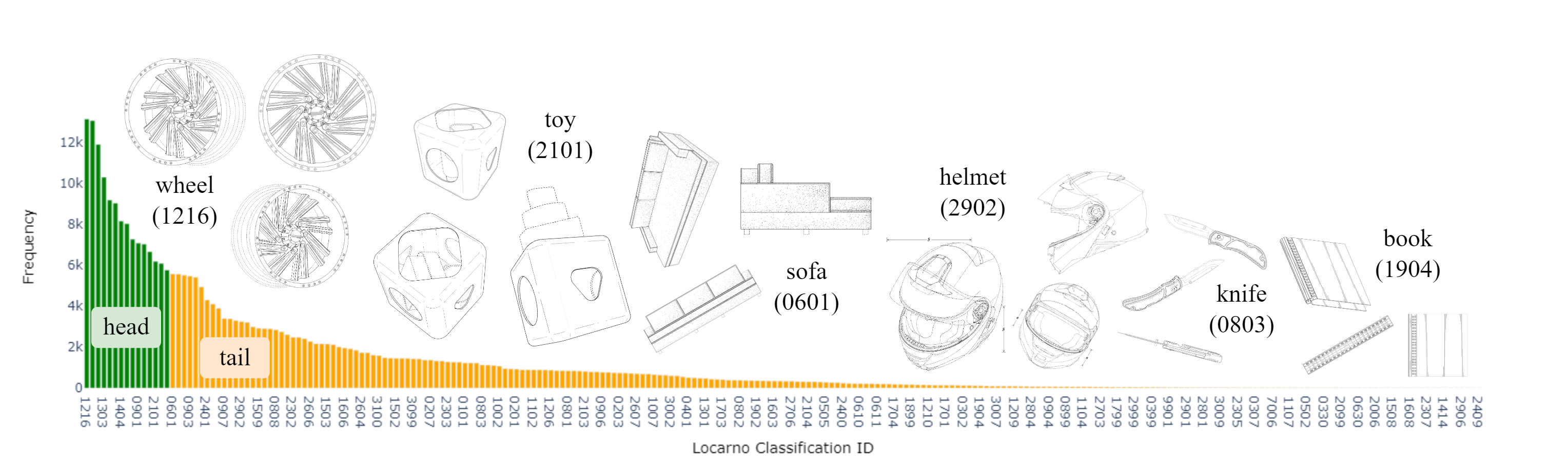}
  \caption{The long-tail distribution of patent images within DeepPatent2 \cite{ajayi2023deeppatent2}, plotting Locarno classification IDs against their frequency. The graph differs between the top 40\% of classes (head) and the bottom 60\% (tail), showing patent image examples from different perspectives with their object names under various classifications.}
  \label{fig:longtail}
\end{teaserfigure}


\maketitle

\section{Introduction}
\label{sec:intro}

Prior art search aims to identify similarities between new inventions and existing technologies, thus ensuring the inventions satisfy novelty and non-obviousness requirements during patent drafting, examination, and infringement analysis \cite{shalaby2019patent}. Traditionally focused on metadata and textual information \cite{krestel2021survey}, researchers have increasingly turned to image-based patent retrieval to overcome the limitations (e.g., the complexity of legal and technical patent language) of textual analysis \cite{jiang2021deriving, kucer2022deeppatent, higuchi2023patent, wang2023learning}, given that patent images provide a clearer, more intuitive understanding of inventions (e.g., vehicle, design, and fashion), enabling faster and deeper insights compared to text alone \cite{carney2002pictorial}.

Patent images, designed to convey technical and scientific information, exhibit distinctive features that set them apart from natural and sketch images. Firstly, they often lack the background context, color, texture, and intensity variability found in natural images, characterized instead by their abstractness and sparseness. Secondly, unlike sketch images, patent images provide detailed and high-quality visualizations from multiple viewpoints. This specificity results in commercial search engines like Google facing difficulties in accurately retrieving relevant patent images from drawing queries \cite{kucer2022deeppatent, wang2023learning}, thereby rendering image-based patent retrieval a significant and ongoing challenge.

Research on image-based patent retrieval, though limited, can be broadly categorized into two: (\textit{i}) Low-level vision-based methods, which employee basic visual features such as visual words \cite{vrochidis2012concept}, shape and contour \cite{tiwari2004patseek, zhiyuan2007outward}, relational skeletons \cite{huet2001relational}, and adaptive hierarchical density histograms \cite{vrochidis2010towards} to describe patent images for retrieval. These methods, however, falter in large-scale applications \cite{wang2023learning}. (\textit{ii}) Learning-based methods have gained traction in recent years. For instance, one early work, using object detection and multi-task framework for patent classification, simultaneously performs image-based retrieval \cite{jiang2021deriving}. With the emergence of DeepPatent dataset \cite{kucer2022deeppatent} and the ECCV 2022 DIRA Workshop Image Retrieval Challenge has prompted exploration into various network architectures, loss functions, and Re-ID techniques to improve retrieval systems \cite{higuchi2023patent, wang2023learning}.

Despite these efforts, past studies have often overlooked the real-world workflow of patent attorneys conducting prior art searches with images. In practice, patent attorneys evaluate not only the visual similarity between current images and those of prior art but also consider the images' descriptions and their associated patent classifications. This oversight leads to several critical gaps and our corresponding contributions: (\textit{i}) Given the importance of textual content, we adopt a visual language model (VLM) \cite{radford2021learning} without following pretrain-finetune paradigm. Furthermore, recognizing the limited semantics in patent images' textual content (i.e., primarily object names and perspectives), we, inspired by past prompting engineering \cite{guo2023viewrefer, zhang2023prompt}, propose using large language models (LLMs) \cite{brown2020language} to generate detailed, alias-containing, free-form descriptions. (\textit{ii}) To incorporate patent classification and address its long-tail distribution (\cref{fig:longtail}), beyond the InfoNCE loss, we introduce multiple coarse-grained losses with uncertainty factors tailored for long-tail data into our VLM. This strategy aims to ensure that patent image representations capture class information while remaining sensitive to the distribution \cite{tian2022vl}. (\textit{iii}) Previous works have treated image-based patent retrieval tasks as Re-ID tasks, which do not fully align with industrial needs. Typically, searches are conducted on large databases and retrieval is carried out both before and after a patent is granted, primarily in two scenarios: novelty detection or prior art search, where a current invention is compared against past inventions to identify similarities; and infringement search, which aims to identify subsequent inventions that might infringe upon the granted patent \cite{krestel2021survey}. Accordingly, we train and validate our model on a larger dataset and ensure that the retrieval metrics align with these temporal concerns. (\textit{iv}) To further understand the practical value of our image-based patent retrieval system, we conducted blind user studies \cite{christensen2014research}. The goal is to directly evaluate and compare the satisfaction, usability, and performance of our method against existing methods in a real-world setting, showing the practical significance of our approach. Hence, we present four-fold distinctive contributions to better meet the industrial demands:
\begin{itemize}
    \item We introduce a language-informed, distribution-aware multimodal approach to patent image feature learning, which is both simple and robust. This method enhances images with corresponding semantic information, augmented via LLMs.
    \item We propose tailored losses specifically designed for the long-tail distribution of patent classifications. This strategy significantly boosts the robustness and accuracy of patent image representations, particularly in scenarios sensitive to class distinctions, leading our method achieve state-of-the-art results.
    \item Our model is validated on a large dataset with metrics specifically tailored for novelty detection, ensuring it meets broad industrial needs.
    \item We employ a multi-paradigm approach, validating the system's effectiveness not only through technical retrieval metrics but also by accentuating its practical utility through user studies.
\end{itemize}

\section{Related Work}
\label{sec:related}

\begin{figure*}[t]
  \centering
  \includegraphics[width=\textwidth]{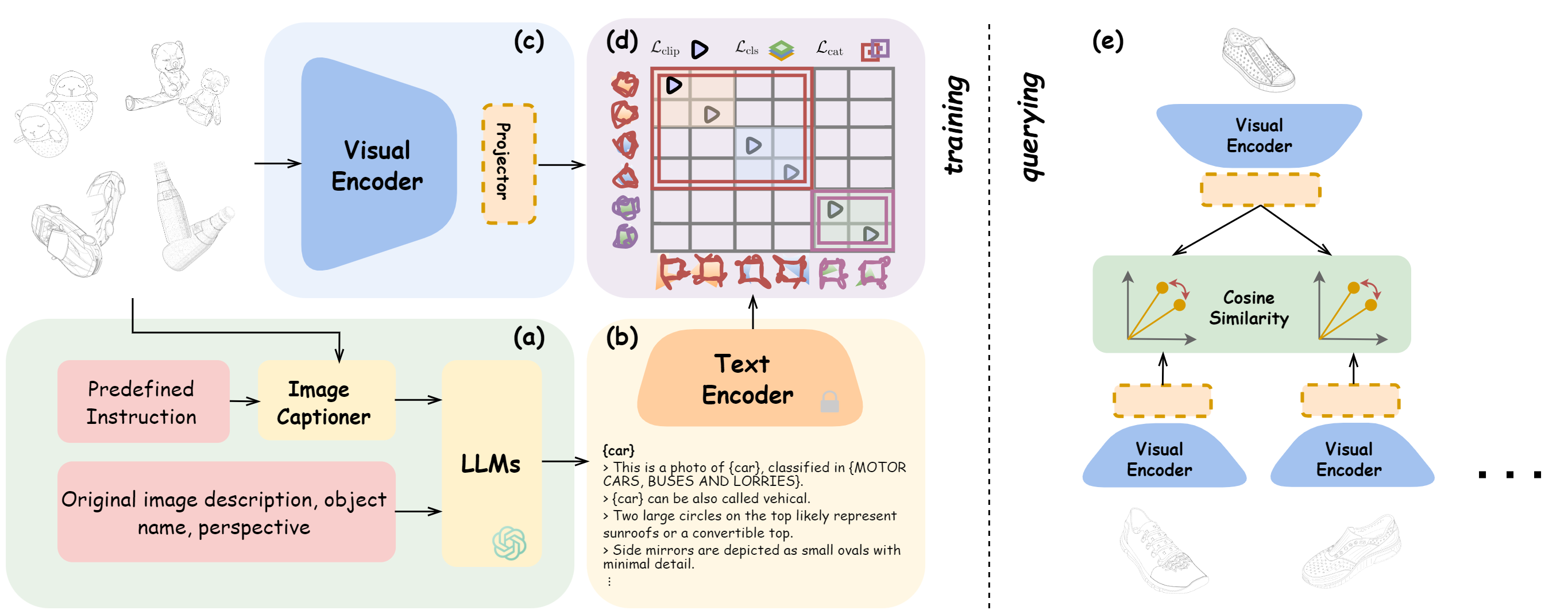}
  \caption{Model Architecture. (a) Generation of diverse, alias-containing, fine-grained descriptions for each patent image using a captioner and LLMs. (b) Text feature extraction from the enriched text via a frozen text encoder. (c) Visual feature extraction through a trainable visual encoder; a projector is employed when a mismatch between text and visual features. (d) Proposed distribution-aware contrastive losses. (e) Query patent images are converted into embeddings for retrieval based on cosine similarity.}
  \label{fig:model}
\end{figure*}

Existing learning-based works on image-based patent retrieval systems can be categorized into two approaches: The first approach is intuitive, starting with the identification of objects within patent images, then training a classifier to associate these identified objects with their respective International Patent Classification (IPC) classes, and extracting vectors from the network for retrieval \cite{jiang2021deriving, vrochidis2012concept, bhattarai2020diagram}. However, this method faces two limitations: (\textit{i}) It relies on objects that the original detector has been pretrained to recognize, resulting in the exclusion of unidentifiable patent images and thus limiting its applicability for large-scale applications. (\textit{ii}) Although it considers IPC, IPC provides a rather coarse classification to the entire patent, which fails to accurately reflect the specific class of a certain image.

The second approach developed with the release of the large-scale DeepPatent dataset \cite{kucer2022deeppatent}, where a series of studies have treated learning patent image representation as a Re-ID (i.e., Patent ID) problem \cite{ye2021deep}. Employing various CNN backbones such as EfficientNet \cite{wang2023learning, higuchi2023patent}, ResNet \cite{kucer2022deeppatent}, ViT \cite{higuchi2023patent}, and SwinTransformer \cite{higuchi2023patent}, these studies aim to embed patent drawings into a common feature space using contrastive loss functions (e.g., triplet loss \cite{kucer2022deeppatent}, ArcFace \cite{higuchi2023patent, wang2023learning}), clustering identical ID images closely and separating different ID images. Although these studies have shown excellent Re-ID capabilities, they have overlooked several critical aspects: (\textit{i}) Re-ID primarily focuses on retrieving images within the same patent, which does not align with the patent industry's workflow (i.e., retrieving images from different patents). (\textit{ii}) The Re-ID approach is susceptible to overfitting, which reduces its generalizability and accuracy in retrieving similar cases across different patents \cite{ma2019large}. (\textit{iii}) These methods often ignore other patent-specific information, such as image descriptions and Locarno classification, which are crucial in practical patent work.

With the rise of VLM and multimodal learning, the retrieval of natural color images has seen significant improvements \cite{liu2021image, karthik2024visionbylanguage, baldrati2023zero, cohen2022my}. Likewise, multimodal methods have become increasingly prevalent in retrieval strategies for sketch images, which are similar, if not identical, to patent images. For example, sketch-based image retrieval involves retrieving natural images using sketch representations \cite{deng2020progressive, sangkloy2016sketchy, xu2022deep, wang2021transferable}. They mainly utilize the associations with natural images to achieve such  effective results. While patent images lack the stroke information found in sketches and are challenging to associate with natural images due to their nature of novelty and multiple perspectives, the textual information and well-defined classification in patents provide a solid foundation for employing a VLM approach. Therefore, we explore the potential of applying VLM to patent image retrieval, an area currently underappreciated, leveraging the rich auxiliary information in patents \cite{ajayi2023deeppatent2}.

\section{Method}
\label{sec:method}

\subsection{Model Overview}
\label{sec:model}

Our objective is to leverage the powerful capabilities of pre-trained LLMs to facilitate feature learning for patent images, thus achieving semantically rich image representations and efficient patent image retrieval. To this end, we introduce a one-stage framework, distinct from the conventional pretrain-finetune approaches employed by prior studies. As depicted in \cref{fig:model}, our model comprises two main components: visual feature extraction and text feature extraction. In the visual component of our model, we preprocess the input patent images using data augmentation techniques tailored for patent images, such as flipping, random cropping \cite{dosovitskiy2020image}, random erasing \cite{zhong2020random}, and gridmask \cite{chen2020gridmask}. We then employ CNN-based backbones to extract visual features from these augmented images. For cases where the output feature dimensions from certain backbones are too long or too short, we utilize projectors composed of MLPs to align the dimensions of visual features with those of text features for subsequent contrastive training (\cref{fig:model} (c)).

In the textual component, our process begins with employing an image captioner, which, given a patent image and a predefined prompt, generates a sophisticated description of the image \cite{liu2024visual, li2023blip, karthik2024visionbylanguage}. This generated description is then combined with other pertinent information about the image, such as its Locarno classification, original image description, object name, and perspective. This composite input is fed into LLMs to produce diversified, alias-caontaining, fine-grained text descriptions, thereby enriching the semantic understanding of the patent image (see \cref{fig:model} (a) and \cref{sec:text}). Following, we employ a frozen text encoder (i.e., text encoder in CLIP), to extract textual features of the descriptions (\cref{fig:model} (b)).

To ensure our contrastive loss is distribution-aware, we introduce three types of loss functions. Firstly, for the conventional VLM contrastive loss $\mathcal{L}_{\text{clip}}$, we treat the pairing of a fine-grained image with its corresponding description as a positive match. For the coarse-grained approach, inspired by \cite{tian2022vl, lo2023enhancing}, we define two scenarios: class-wise, where an image and a sentence from the same \textit{class} are considered a positive pair ($\mathcal{L}_{\text{cls}}$); and category-wise, where an image and a sentence from the same \textit{category} (e.g., head or tail categories) are seen as a positive pair ($\mathcal{L}_{\text{cat}}$). These losses are combined to update the visual encoder during the training phase (see \cref{fig:model} (d) and \cref{sec:distribution}).

In the query phase (\cref{fig:model} (e)), each patent image is transformed into embeddings with the visual encoder and then stored in a vector database. Retrieval of other images is done by comparing these embeddings via cosine similarity, where embeddings closer in distance are ranked higher, and those farther apart are ranked lower.

\subsection{Distribution-Aware Contastive Loss}
\label{sec:distribution}

As previously mentioned, we formulate our task as a VLM contrastive learning paradigm. The traditional instance-based training objective for a single image can be described as follows (i.e., InfoNCE loss, \cref{eq:instance}):
\begin{equation}
  \mathcal{L}_{\text{clip}} = -\log(\dfrac{\exp(\mathbf{t}_+^T \cdot \mathbf{v}/ \tau)}{\sum_{i=1}^B \exp(\mathbf{t}_i^T \cdot \mathbf{v}/ \tau)}),
  \label{eq:instance}
\end{equation}
where $(\mathbf{t}_1^T, \mathbf{t}_2^T, ..., \mathbf{t}_B^T)$ denotes $B$ text features extracted by the text encoder and $\mathbf{v}$ denotes the learned feature of a patent image. The term $\mathbf{t}^T \cdot \mathbf{v}$ represents the cosine similarity score between the patent image and texts and $\tau$ is a learnable temperature coefficient. This objective is to maximize $\mathbf{t}_+^T \cdot \mathbf{v}$, which indicates the feature similarity between the patent image and the corresponding textual information. 

However, relying solely on instance-based contrastive loss falls short in capturing class or category information, potentially leading to suboptimal performance in datasets with skewed distributions \cite{lo2023enhancing}. Therefore, we propose \textit{class}-based and \textit{category}-based coarse-grained losses (i.e., $\mathcal{L}_{\text{cls}}$ and $\mathcal{L}_{\text{cat}}$, see \cref{fig:model} (d)). These losses can be described in a similar form as follows (\cref{eq:loss}):
\begin{equation}
\begin{split}
   &- \dfrac{1}{|\mathbf{V}_i^+|}\sum_{\mathbf{V}_j \in \mathbf{V}_i^+} \log\dfrac{\exp(\mathbf{t}_i^T \cdot \mathbf{v}_j/ \tau)}{\sum_{\mathbf{V}_k \in \mathbf{V}} \exp(\mathbf{t}_i^T \cdot \mathbf{v}_k/ \tau)} \\
   &- \dfrac{1}{|\mathbf{T}_i^+|} \sum_{\mathbf{T}_j \in \mathbf{T}_i^+} \log\dfrac{\exp(\mathbf{t}_j^T \cdot \mathbf{v}_i/ \tau)}{\sum_{\mathbf{T}_k \in \mathbf{T}} \exp(\mathbf{t}_k^T \cdot \mathbf{v}_j/ \tau)}, 
   \end{split}
  \label{eq:loss}
\end{equation}
where $V$ represents a batch of patent images, and $T$ denotes the corresponding set of text sentences. The subset $T_i^+$ comprises texts that share the same class (in the case of $\mathcal{L}_{\text{cls}}$) or category (for $\mathcal{L}_{\text{cat}}$) with the image $V_i$. Similarly, $V_i^+$ includes all images that share the same class or category with the text $T_i$. By doing so, our model gains an understanding of the class and category information, enabling it to acquire robust representations even for those in the tail classes. Furthermore, since the text description for each image sample varies with each iteration, and combined with the class or category loss, the one-to-one pairing relationship between images and texts is less rigid. This variability acts as an additional regularization mechanism, preventing the model from adhering to fixed, trivial correlations within specific image-text pairs.

Considering each loss's stability varies, we move away from linear loss combination towards a method based on homoscedastic uncertainty \cite{kendall2017uncertainties, kendall2017geometric}, learnable through probabilistic deep learning. This type of uncertainty, independent of input data, reflects the task's intrinsic uncertainty. The loss includes residual regression and uncertainty regularization components. The implicitly learned variance $\hat{s}$ moderates the residual regression, while regularization prevents the network from predicting infinite uncertainty. Hence, The overall loss can be written as in \cref{eq:overall}.
\begin{equation}
\begin{split}
  \mathcal{L} = &\mathcal{L}_{\text{clip}}\exp(-\hat{s}_{\text{clip}}) + \hat{s}_{\text{clip}} \\
  &+ \mathcal{L}_{\text{cls}}\exp(-\hat{s}_{\text{cls}}) + \hat{s}_{\text{cls}} \\
  &+ \mathcal{L}_{\text{cat}}\exp(-\hat{s}_{\text{cat}}) + \hat{s}_{\text{cat}}
  \end{split}
  \label{eq:overall}
\end{equation}
where $\hat{s}$ is learnable homoscedastic uncertainty. We find this loss is robust to our task.

\begin{table*}[t]
\caption{Evaluation results across models using metrics such as mAP, Recall@K, and MRR@K, under different settings. Highest values are highlighted in bold.}
\label{tab:results}
\centering
\resizebox{\textwidth}{!}{
\begin{tabular}{llrrrrrrrrrrrr}
\toprule
                                 &                        & \multicolumn{4}{c}{Head Classes}                 & \multicolumn{4}{c}{Tail Classes}                 & \multicolumn{4}{c}{All Classes}                  \\
\cmidrule(lr){3-6} \cmidrule(lr){7-10} \cmidrule(lr){11-14}
                                 &                        & mAP & R@5 & R@10 & M@10 & mAP & R@5 & R@10 & M@10 & mAP & R@5 & R@10 & M@10 \\
\midrule
\multirow{3}{*}{Baseline}        & PatentNet \cite{kucer2022deeppatent}             & 22.1 & 13.4 & 26.2 & 35.7 & 12.9 & 3.1  & 11.6 & 25.2 & 15.8 & 8.0  & 16.8 & 29.4 \\
                                 & EfficientNetB0+ArcFace \cite{wang2023learning} & 31.5 & 19.8 & 34.9 & 43.9 & 20.1 & 8.7  & 20.3 & 33.1 & 24.1 & 13.6 & 26.7 & 37.4 \\
                                 & SwinV2-B+ArcFace \cite{higuchi2023patent}      & 32.0 & 19.4 & 34.2 & 45.2 & 20.3 & 6.3  & 21.4 & 32.7 & 25.1 & 11.6 & 27.3 & 37.7 \\
\midrule
\multirow{4}{*}{\textbf{Ours} (backbone)} & ResNet50               & 31.9 & 35.1 & 48.9 & 45.7 & 22.4 & 20.9 & 38.2 & 36.3 & 25.6 & 26.6 & 42.2 & 40.1 \\
                                 & EfficientNetB-0        & 64.2 & 43.5 & 58.5 & 76.4 & 43.0 & 31.2 & 44.0 & 55.7 & 52.4 & 36.9 & 50.5 & 63.1 \\
                                 & ViT-B-32               & \textbf{78.0} & 51.7 & 65.3 & 90.3 & \textbf{61.6} & 36.7 & \textbf{53.5} & \textbf{75.4} & \textbf{69.1} & 43.4 & \textbf{58.6} & \textbf{81.3} \\
                                 & SwinV2-B               & 77.2 & \textbf{52.5} & \textbf{66.2} & \textbf{90.8} & 61.3 & \textbf{42.5} & 52.5 & 75.0 & 67.8 & \textbf{46.0} & 57.7 & 80.6 \\
\bottomrule
\end{tabular}
}
\end{table*}

\subsection{Text Enrichment}
\label{sec:text}

Converting object names, perspectives, class names, and the patent image's original descriptions into text for input into a text encoder is a plausible way to generate text embeddings for supervisory purposes. However, this method encounters several limitations. Firstly, the most comprehensive descriptions provided by patents are typically succinct and straightforward, such as \textit{FIG. 3 is a front elevational view of the light device}, leading to unclear and sparse information about the image. Additionally, the similarity and potential overlap among different classes (e.g., \textit{automobiles}, \textit{motor cars}, and \textit{toy cars}) can obscure the distinction of nuanced concepts. To overcome these challenges and derive more discriminative text features, we employee captioners \cite{liu2024visual} and LLMs \cite{brown2020language} for producing detailed, enriched descriptions that enhance the semantic understanding of the images.

Specifically, we firstly employ captioners to generate descriptions of images in a manner that would capture aspects patent attorneys focus on. We provide the captioners with images alongside a set of predefined instructions that guide the description process. For example, instructions might include: \textit{Describe the distinct visual elements present in the design, such as shapes, contours, texture, and the arrangement of various components.}  The outcomes of this process, merged with pre-existing auxiliary information and predetermined instruction templates, are then fed into LLMs. For example, we employ templates such as \textit{This is a photo of} \texttt{\{Object Name\}}, \textit{classified as} \texttt{\{Class\}}, \textit{This image features} \texttt{\{Details\}}, and \texttt{\{Object Name\}} \textit{can also be referred to as} \texttt{\{Synonym\}}, to generate enriched text. Ultimately, this approach yields around 20 detailed text descriptions per image, designed to mine the semantic nuances within the text feature space.

Additionally, our research revealed that utilizing more specific object names significantly enhances feature learning. For example, the broad class of \textit{Emergency equipment}, which encompasses distinct items such as \textit{lighting fixture}, \textit{horticulture grow light}, and \textit{lighting device}. Therefore, in our text generation process, we prioritize these detailed object names over the more generic class names, different from the previous approach that typically leans on class categorization \cite{rozenberszki2022language, huang2024joint}.

\subsection{Metrics for Patent Retrieval}
\label{sec:metrics}

As mentioned before, in image-based patent retrieval, particularly for novelty detection and related work searches, patent professionals explore databases of patents, applications, and scientific literature to determine an invention's uniqueness. This process, crucial both before and after a patent application is filed, focuses on identifying if similar prior inventions exist. Accordingly, when evaluating retrieval metrics, it's necessary to account for the temporal factor, ensuring that only prior art—rather than contemporaneous or subsequent inventions—is considered for retrieval \cite{krestel2021survey}. Considering a database $D$ where each data point is represented by a tuple $(\mathbf{v}, t)$, with $\mathbf{v}$ being the image embedding and $t$ the granted time of the patent the image belongs to. For each query image $\mathbf{v}_q$ with a granted time $t_q$, define $D'_q \subseteq D$ containing images granted before $t_q$. Hence, the following retrieval metrics should be calculated over $D'_q$ given $\mathbf{v}_q$: (\textit{i}) We use mean Average Precision (mAP), a metric obtained by averaging AP scores across all classes. (\textit{ii}) Following previous works, the standard evaluation protocol \cite{baldrati2022effective} is to report the recall at rank $K$ (Recall @ $K$ or R$@K$) at different ranks ($5, 10$). (\textit{iii}) We calculate the Mean Reciprocal Rank @ $K$ with temporal concern (MRR$@K$ or M$@K$), which averages the reciprocal of the rank for the first correctly predicted patent image within the top 10 rankings across all test samples.

\section{Experiments}
\label{sec:experiments}

\subsection{Implementation Details}
\label{sec:implementation}

In our experiments, we employed PyTorch \cite{paszke2019pytorch} and utilized clusters of NVIDIA A100 GPUs. For the VLM, we explored various ViT variants \cite{dosovitskiy2020image}, ResNet50 \cite{he2016deep}, EfficientNetB-0 \cite{tan2019efficientnet}, and SwinV2-B \cite{liu2021swin, liu2022swin} as backbones for the visual encoder. The text encoder was adapted from the original CLIP model \cite{radford2021learning}, remaining fixed throughout the experiments. For captioner, we leveraged open-source BLIP-2 \cite{li2023blip} and GPT-4V \cite{achiam2023gpt}. Regarding LLMs, our focus was on GPT-4 \cite{achiam2023gpt}, though we also experimented with other LLMs like GPT-3.5-Turbo \cite{brown2020language} and LLaMA-2 \cite{touvron2023llama}.

For our experiments, we employed the DeepPatent2 dataset. Previous research has mainly adopted the original DeepPatent dataset \cite{kucer2022deeppatent}; however, this dataset suffers from a narrow collection span, lacks image-related metadata, and does not segment sub-images. These limitations can introduce substantial noise in inter-image relationships. Fortunately, the DeepPatent2 \cite{ajayi2023deeppatent2} dataset addresses these issues effectively. To maintain comparability with previous methods, we utilized DeepPatent2 data from 2016 to 2019, consisting of 822,792 records with 407 Locarno classes. Of these, 90\% were used for the training set, and 10\% for the validation set. For our query dataset, we used 252,296 records from the year 2020.

Our baseline models are our replicating state-of-the-arts on the same dataset to ensure comparability: these include PatentNet \cite{kucer2022deeppatent}, SwinV2-B+ArcFace \cite{higuchi2023patent}, and EfficientNet+ArcFace \cite{wang2023learning}. Our primary experiments focused on a variety of visual encoder backbones, such as ResNet50, EfficientNetB-0, ViT-B-32, and SwinV2-B. Due to space constraints in this paper, detailed results from experiments involving the captioner and LLMs will be presented in the full manuscript. For preliminary insights, we utilized GPT-4 for both the captioner and LLMs. In our ablation study, we explore the effects of distribution awareness losses (i.e., $\mathcal{L}_{\text{cls}}$ and $\mathcal{L}_{\text{cat}}$), the text generation component, and the captioner.

\begin{table}[h]
\caption{Evaluation results of an ablation study on various components within the model architecture.}
\label{tab:ablation}
\centering
\resizebox{\linewidth}{!}{
\begin{tabular}{lrrrrrr}
\toprule
                & \multicolumn{2}{c}{Head Classes} & \multicolumn{2}{c}{Tail Classes} & \multicolumn{2}{c}{All Classes} \\
\cmidrule(lr){2-3} \cmidrule(lr){4-5} \cmidrule(lr){6-7}
                & mAP & R@10 & mAP & R@10 & mAP & R@10 \\
\midrule
Baseline        & 32.3 & 34.7 & 21.9 & 30.2 & 26.0 & 32.8 \\
$\mathcal{L}_{\text{cls}}$ and $\mathcal{L}_{\text{cat}}$ & 47.4 & 49.2 & 47.0 & 48.4 & 47.9 & 49.7 \\
Text Generation & 70.2 & 59.7 & 52.4 & 53.6 & 58.7 & 56.3 \\
Captioner       & 78.0 & 65.3 & 61.6 & 53.5 & 69.1 & 58.6 \\
\bottomrule
\end{tabular}
}
\end{table}

\begin{figure*}[]
  \centering
  \includegraphics[width=\textwidth]{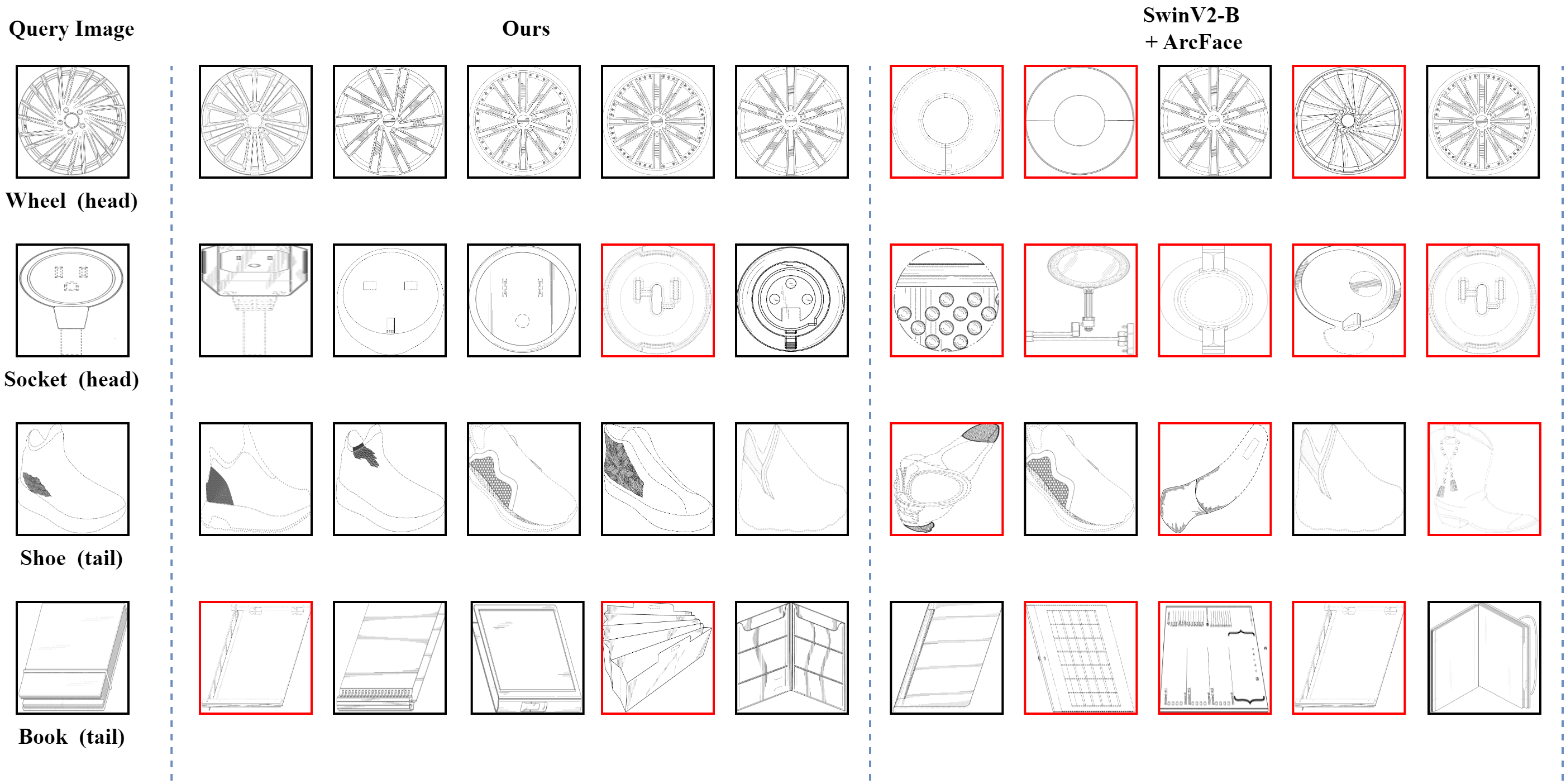}
  \caption{Qualitative Results of the Image-Based Patent Image Retrieval System. The leftmost image represents the query image, annotated with its object name and classification as either a head or tail class. The middle section displays the retrieval results from our method, where images framed in black indicate a match with the query image's class, and those framed in red indicate a mismatch. The right section shows the comparative results using the previous state-of-the-art method (SwinV2-B + ArcFace).}
  \label{fig:quali}
\end{figure*}

\subsection{Experimental Results}
\label{sec:experimental}

Table \ref{tab:results} presents the quantitative results on the DeepPatent2 query set. Overall, our approach significantly outperforms the state-of-the-art, achieving up to a 53\% improvement in the mAP metric, 38\% in Recall @ 5, 41.8\% in Recall @ 10, and 51.9\% in MRR @ 10. Notably, both ViT-B-32 and SwinV2-B show comparable performance, with each excelling under different metrics or scenarios. For instance, ViT-B-32 performs better in tail classes, while SwinV2-B shows strength in head classes, suggesting an interaction between data distribution and model architecture. With these results, we have achieved the state-of-the-art in this task. Given the strong performance across all classes with ViT as the backbone, we will base our ablation study on this model to further investigate the impact of various model components on performance.

Table \ref{tab:ablation} presents the evaluation results of an ablation study on four components, starting with a baseline model, which is a standard CLIP model. The subsequent rows represent enhancements to this baseline. Firstly, by incorporating a distribution-aware contrastive loss, we observe significant performance gains in both head and tail classes, with tail classes experiencing more substantial improvements (approximately 20\% in mAP and Recall @ 10), and head classes seeing a 10\% increase in these metrics. Next, by incorporating text generation functionality, which is guided by the LLM, the model learns richer semantic relationships between images. The addition of this feature leads to a 10-20\% improvement across various metrics. Finally, by integrating a captioner module, the key details of the design inventions in the images are directly expressed by the captioner, further highlighting the semantic focal points of the images and enhancing the retrieval performance.

\subsection{Qualitative Results}
\label{sec:qualitative}

Qualitative results (see \cref{fig:quali}) indicate that our approach retrieves better results than the previous state-of-the-art, evident in both head and tail classes. Focusing solely on our model, it is apparent that it underperforms in tail classes. Additionally, our system not only retrieves the correct class given an image but also finds images that are visually similar to the query image. Furthermore, we delve into the errors made by the previous state-of-the-art. For example, when presented with an image labeled \textit{shoe}, the previous model might retrieve a shoe image, but it actually belongs to the category of \textit{shoelaces}. This reveals that the previous approach did not align semantic information within the images, often leading to the retrieval of visually similar but categorically different images. Similar issues occur with categories like \textit{vehicles \& toy cars} or \textit{flashlights \& chargers}, where the images look alike but differ semantically. Our model mitigates these errors by guiding the image's embedding space with linguistic information, which enhances semantic alignment.

\begin{figure}[]
  \centering
  \includegraphics[width=\linewidth]{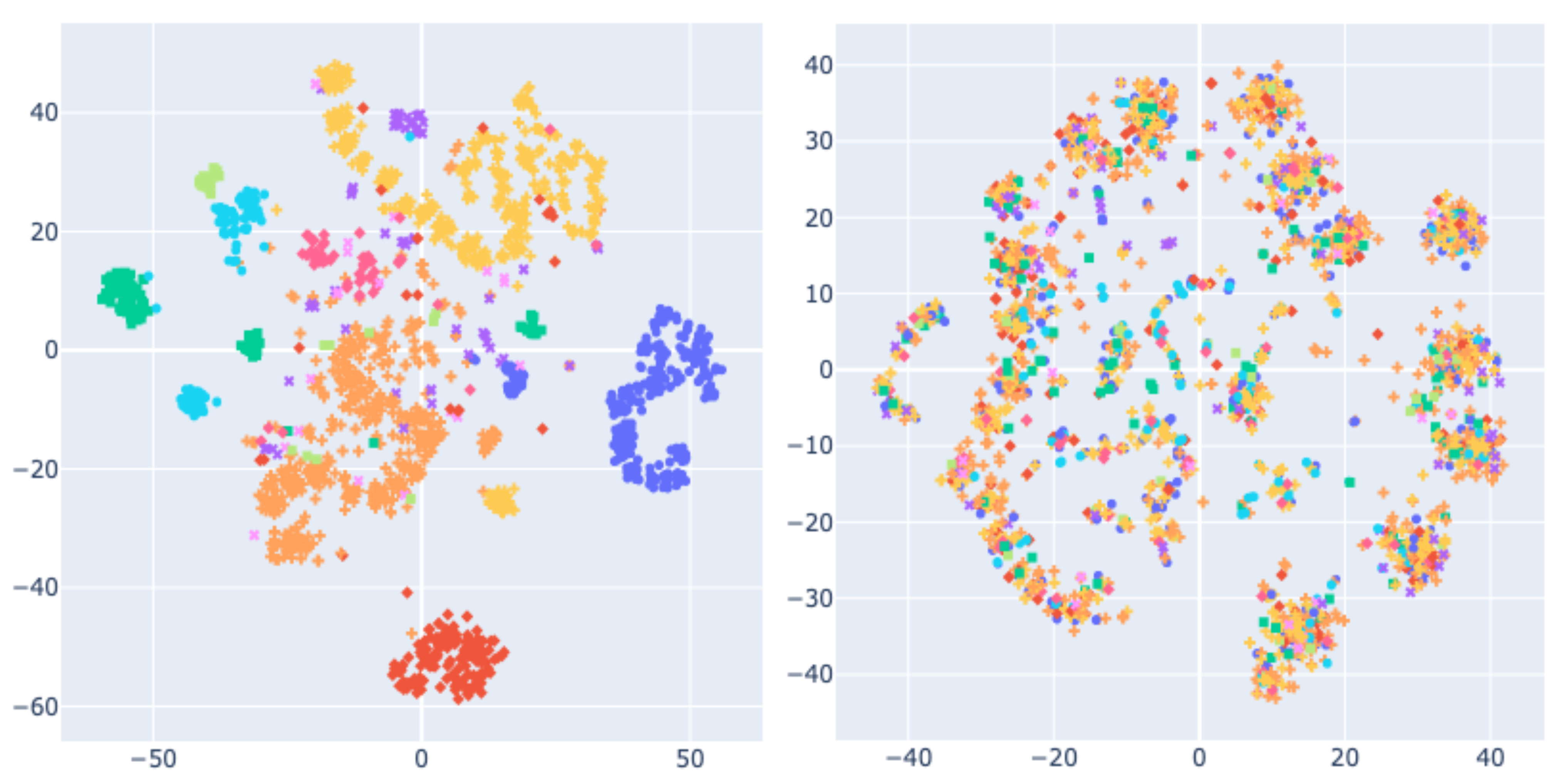}
  \caption{t-SNE Visualization of Image Embeddings. This figure presents a two-dimensional t-SNE projection of randomly sampled 2,000 image embeddings, with each axis representing one dimension. Different colors and shapes in the plot indicate distinct classes. The left subplot illustrates the results from our model, while the right subplot displays the results using the previous state-of-the-art, SwinV2-B+ArcFace.}
  \label{fig:tsne}
\end{figure}

Based on the t-SNE results shown in \cref{fig:tsne}, our method yields more clustered embeddings, suggesting that the model effectively captures the inherent structures or classes within the data. Each cluster represents a group of similar images, closely corresponding to predefined classes, indicating that the model has learned meaningful and discriminative features for each class. This clustering enables the model to effectively distinguish between different classes.

Conversely, the previous approach results in a t-SNE visualization with less clustered and more continuous embeddings, making it difficult to identify distinct clusters or their correlation to predefined classes. This indicates that the model's learned representations are less discriminative, potentially capturing more generalized features shared across multiple classes. Although some clustering is visible, it may not relate to actual classes but rather to visually similar images, blurring the boundaries between different classes.

\section{User Study}
\label{sec:user}
To further ensure the practical value of our system, we conducted a user study following rigorous psychological procedures. As for participants, we recruited 15 patent agents (48\% female, average age: 33.4 years) to perform tasks related to design patent image retrieval.

As for procedure, we employed a double-blind test, where participants were unaware of the underlying retrieval system during their tasks. They could encounter either our retrieval system or a system based on the previous approach. Each patent agent handled 30 retrieval tasks, with these tasks randomly assigned to one of the two systems—15 tasks with our system and 15 with the previous approach. After each task, participants rated their satisfaction with the retrieval results on a scale from 1 to 5 and recorded the time taken to complete the task (in hours). To minimize randomness, we averaged the scores across systems for each participant, resulting in four scores per person (two systems $\times$ two scores).

The results of the paired t-test revealed significant differences in satisfaction levels, with patent agents showing a higher satisfaction with our system compared to the previous approach, $t$(14) = 3.30, $p < 0.01$. Regarding task completion time, agents completed tasks faster using our system, $t$(14) = -4.30, $p < 0.001$. These results indicate that our system is more efficient and better meets the practical needs of professionals in the field.

\section{Conclusion}
\label{sec:conclusion}

Our method has achieved new state-of-the-art results in the quantitative evaluation of mAP, Recall@$K$, and MRR@$K$, as well as in high-quality image retrieval during qualitative evaluation. For many years, current commercial design patent retrieval systems have had significant shortcomings. For example, traditional text-based searches can be limiting due to the subjective interpretation of design features and the difficulty in describing visual details with text. Although learning-based image retrieval systems have started to emerge in the last two years, their practical value remains limited.

To address these issues, our proposal can effectively solve this problem. Firstly, we proposed a new learning-based architecture capable of learning image features with practical value. These representations not only contain visual information but are also aligned with corresponding (augmented) semantic text and classification data. This has substantial practical value because specific graphic semantic features such as curvature, edges, and geometric details are considered. Focusing solely on the image itself might overlook these critical visual elements. Secondly, we utilize a larger and more well-defined dataset, which encompasses a broader collection span, image-related metadata, and segmented sub-images. This makes the model more robust and enhances its accuracy. Thirdly, addressing the long-tail distribution in classification, our study is the first to propose distribution-aware losses, which have proven to be effective. Lastly, we conducted a user study to demonstrate the practical value of our system in the field, showing that it is more efficient, accurate, and time-saving.

In the future, we have several directions for further expansion: ($i$) Identifying similarities between this invention and prior arts, which involves not only using existing models for visualizations through explainable AI \cite{xu2019explainable} but also leveraging data from examiner's reports (Office Actions) to guide further explorations. ($ii$) While our research currently focuses on prior art searches, future work could also explore other temporal dimensions, such as infringement searches. Additionally, image domain adaptation \cite{wang2018deep} could be used to enhance the effectiveness of searches across different domains, such as retrieving E-commerce images using patent drawings.


\bibliographystyle{ACM-Reference-Format}
\bibliography{sample-base}










\end{document}